\documentclass{article}
\usepackage{spconf,amsmath,graphicx}
\usepackage{url}
\usepackage{booktabs}
\usepackage{CJKutf8}
\usepackage{makecell}
\usepackage{cite}
\usepackage{hyperref}
\usepackage{xcolor,colortbl}
\definecolor{Gray}{gray}{0.95}
\newcolumntype{C}{>{\columncolor{Gray}}c}
\usepackage{color}
\usepackage{cite}
\usepackage{amsfonts}       
\usepackage{xcolor}         
\usepackage{graphicx}
\usepackage{amsmath,amssymb}
\usepackage{xurl}
\usepackage{multirow}
\usepackage{tabularx}
\usepackage{booktabs}
\usepackage{makecell}
\usepackage{float}
\hypersetup{colorlinks=true}
\usepackage{caption}
\usepackage{subcaption}
\usepackage{cite}
\usepackage{tabularx}
\usepackage{booktabs}
\usepackage{adjustbox}
\usepackage{amsmath}
\usepackage{svg}

\title{Do Prompts Really Prompt? \\ Exploring the Prompt Understanding Capability of Whisper}
\name{Chih-Kai Yang$^1$, Kuan-Po Huang$^{1,2}$, Hung-yi Lee$^1$}
\address{$^1$National Taiwan University \quad $^2$ASUS Intelligent Cloud Services}
\begin{document}
\ninept
\maketitle
\begin{abstract}
This research explores how the information of prompts interacts with the high-performing speech recognition model, Whisper. We compare its performances when prompted by prompts with correct information and those corrupted with incorrect information. Our results unexpectedly show that Whisper may not understand the textual prompts in a human-expected way. Additionally, we find that performance improvement is not guaranteed even with stronger adherence to the topic information in textual prompts. It is also noted that English prompts generally outperform Mandarin ones on datasets of both languages, likely due to differences in training data distributions for these languages despite the mismatch with pre-training scenarios. Conversely, we discover that Whisper exhibits awareness of misleading information in language tokens by ignoring incorrect language tokens and focusing on the correct ones. In sum, We raise insightful questions about Whisper's prompt understanding and reveal its counter-intuitive behaviors. We encourage further studies.
\end{abstract}
\vspace{-5pt}
\section{Introduction}
\label{introduction}
The rise of large-scale foundation models in diverse fields, such as natural language processing (NLP)~\cite{flan, t0, llama2, vicuna2023, openai2024gpt4technicalreport} and computer vision~\cite{interimage, zhang2024visionlanguage, cogvlm, shi2024non, luo2024fairclip, lin2024vilapretrainingvisuallanguage}, has ignited interest in various studies and applications. Among these, prompting, a technique aimed at unleashing the potential of foundation models~\cite{cot, chen2023unleashing, li2023large, gu2023systematic, li2022making, zhang2023multimodal},  has gained prominence in the AI community. It has also been extended to speech processing, like text-to-speech 
~\cite{guo2023prompttts, kharitonov2023speak} and spoken language understanding~\cite{wavprompt}.

Notably, Whisper~\cite{whisper}, a cutting-edge automatic speech recognition (ASR) model, becomes a robust backbone for speech-processing systems~\cite{whislu, rathod23_interspeech, gong23d_interspeech, chen2024does, ameer2023whisper, ning2023vitsbased, christ2024papersmartestreviewers, zhang2024improvingchildspeechrecognition, ameer2024optimizing, le2024phowhisper, cui2024spontaneousspeechbasedsuiciderisk} and has shown promise in prompting as prior works showed that prompting, using textual prompts or special tokens, improved Whisper's performances in scenarios like code-switched (CS) ASR~\cite{concat, yang2023investigating}, audio-visual ASR~\cite{concat}, etc. 

Typically, these works assume that Whisper can understand and capture useful information from prompts. However, there is a domain shift between its pre-training and prompting scenarios. Take prompting with textual prompts~\cite{concat, yang2023investigating, wang2024can, lyricwhiz} as an example. During pre-training, Whisper predicts the next token based on the transcription of the preceding speech segment, which is semantically related to the input. However, in the prompting applications, the previous context is replaced with custom prompts that are not necessarily related to the content of the input speech. This mismatch raises natural questions about Whisper's ability to understand those custom prompts. Similarly, domain shifts exist when prompts are in different languages from the speech data, or when Whisper's special tokens are manipulated, as these scenarios contradict the pre-training setting.


Research in NLP~\cite{webson-pavlick-2022-prompt} has shown that models can be enhanced by prompting even without fully understanding the prompts, suggesting that performance improvements do not necessarily imply true prompt comprehension because prompts with incorrect information may also improve the model like those with correct information, which is not expected for a model with true prompt understanding. Hence, in light of the domain shifts mentioned above, the assumption that Whisper understands prompts requires scrutiny despite the observed performance gains in earlier studies. Additionally, a deeper understanding of the prompting mechanism provides better insights into Whisper, facilitating further development and application.


This work explores Whisper's prompt understanding by analyzing its performances when prompted with information that matches or mismatches with the testing data in several scenarios. The information is encoded into prompts through special tokens or certain textual templates, and thus the ``prompt understanding" is required for the model to process the information. Ideally, if the model possesses good prompt understanding, matched prompts should lead to better performance. Our systematic analysis based on three novel metrics reveals that Whisper exhibits limited semantic understanding of textual prompts, often performing better with mismatched information. We also find that textual prompts hurt the performance in general. Further analysis surprisingly reveals that no positive correlation between prompt understanding and performance exists, showing that prompting Whisper does not work as expected. Additionally, prompts in different languages from the testing data often outperform those in the same language, showcasing a counter-intuitive effect of prompt language. In code-switched ASR, expected performance degradation occurs with nonexistent language token pairs, showing that Whisper can filter out irrelevant information within the language tokens, likely benefiting from extensive pre-training.



Overall, our contributions\footnote{Resources of this work can be found at \url{https://github.com/b08202033/whisper_prompting}.} are: 1) Pioneering exploration of Whisper's prompt understanding ability with three newly proposed metrics, 2) Uncovering the unexpected behaviors of Whisper when given textual prompts and empirically demonstrating that adhering strongly to topic information in prompts does not guarantee improved performances, 3) Revealing a counter-intuitive effect of prompt languages where English prompts outperform Mandarin ones, even with Mandarin testing data, and 4) Highlighting its ability to disregard misleading information in language tokens. These findings serve as solid foundations for further research and applications on prompting methods and the development of Whisper and other universal ASR models with strong prompt understanding.


\section{Related works}
\label{relatedwork}
\subsection{Whisper}
Whisper is a family of encoder-decoder ASR models. The input of the decoder includes special tokens, e.g., language tokens like $<$$|\mathtt{en}|$$>$, task tokens like $<$$|\mathtt{transcribe}|$$>$, etc., to convey specific information. Specifically, the $<$$|\mathtt{prev}|$$>$ token signifies the transcript of the previous utterance in long-form transcription. Prompting is achieved by replacing previous contexts with custom prompts, termed \textit{textual} prompts. Following Peng et al.~\cite{concat}, the concept of prompts is extended to special tokens. Complete prompts resemble $<$$|\mathtt{prev}|$$>$ $\mathtt{textual}$ $\mathtt{prompt}$ $<$$|\mathtt{startoftranscript}|$$>$ $<$$|\mathtt{language}|$$>$ $<$$|\mathtt{transcribe}|$$>$.

\subsection{Prompting Whisper}
Prompting Whisper is widely applied. It has been known that prompting can make Whisper follow a certain style~\cite{cookbook} when transcribing, e.g., transcribing in Traditional Chinese instead of Simplified Chinese when given a prompt in Traditional Chinese, etc. However, the application of prompting Whisper is much more than this. Peng et al.~\cite{concat} showed that Whisper can be prompted for audio-visual ASR by including visual objects retrieved by CLIP~\cite{clip} in textual prompts. They also showed that concatenating two language tokens enhanced Whisper on code-switched (CS) ASR for those languages. Yang et al.~\cite{yang2023investigating} observed improvement in CS ASR by indicating code-switching occurrences in textual prompts.  Wang et al.~\cite{wang2024can} showcased the efficacy of task-informed textual prompts, e.g., ``recognize dialect speech", and in-context learning in boosting Whisper on dialect ASR. Zhuo et al.~\cite{lyricwhiz} generalized Whisper to lyric transcription via prompting. These studies show the potential of prompting Whisper beyond style imitation. However, whether Whisper understands the prompts remains unexplored, motivating our systematic investigation of Whisper's prompt understanding.

\subsection{Assessing prompt understanding}
Prompt understanding is a crucial research topic in NLP~\cite{webson-pavlick-2022-prompt, min-etal-2022-rethinking, shivagunde-etal-2024-deconstructing-context}. It is often assessed by comparing the model's performances when given prompts that correctly provide relevant and meaningful task information versus corrupted prompts. Corruptions include misleading prompt templates for natural language inference (NLI) that instruct the model to perform unrelated tasks~\cite{webson-pavlick-2022-prompt}, wrong labels for in-context learning examples~\cite{min-etal-2022-rethinking}, and random word insertions in the prompts~\cite{shivagunde-etal-2024-deconstructing-context}. As these corruptions provide incorrect information, a model with good prompt understanding should perform worse with corrupted prompts. This approach allows for a quantitative assessment of prompt understanding. However, it has been found that corruption sometimes brings improvements rather than degradation to language models, suggesting performance improvement does not necessarily indicate prompt understanding. In this paper, we assess Whisper's prompt understanding with a similar comparative methodology despite the improvements reported in prior works.

\section{Prompt understanding of Whisper}
\subsection{Problem formulation}
\label{problemformulation}
This work aims to rethink the prompt understanding capabilities of Whisper. We mainly focus on the following questions:
\begin{enumerate}

    \item \textbf{Semantic understanding of textual prompts}: Does it understand and capture useful semantic information from the textual prompts for transcribing (Sec. \ref{textual})?

    \item \textbf{Effect of textual prompt languages}: Does it benefit from the implicit language information when prompts and testing data are in the same language (Sec. \ref{prompt_language})?

    \item \textbf{Encoded information in language tokens}: Does it understand the language information in language tokens (Sec. \ref{language_token})?
\end{enumerate}

These questions address essential aspects of prompting methods for Whisper, grounded in existing research and observations, and warrant thorough investigation. The first question is motivated by the prior works that use textual prompts to enhance Whisper's performances on various tasks~\cite{concat, yang2023investigating, wang2024can, lyricwhiz} and the mismatch between prompting applications and pre-training settings. This discrepancy necessitates revisiting Whisper's understanding of textual prompts.

The second question examines whether prompts in the same language as the testing speech are more beneficial than prompts in other languages. During pre-training, Whisper predicts the next tokens conditioned on the previous context, likely in the same language as the currently decoded speech. Thus, it is reasonable to hypothesize that when replacing the previous context with custom prompts, prompts in the same language as the testing data should be more helpful, as they align with pre-training settings and may implicitly convey the language information. Prior observations that Whisper can follow the style of textual prompts support this hypothesis since language can be viewed as a style. We investigate this hypothesis.

Finally, in light of prior works~\cite{concat, aditya2024attention} using language tokens and associated attention weights to improve Whisper on multilingual and code-switching tasks, we raise the third question and explore if Whisper understands the encoded information in language tokens. 

These questions address Whisper's semantic and language-related understanding of textual prompts and language tokens, providing valuable insights for further research and applications toward developing a universal ASR model with true prompt understanding.



\subsection{Method}
\label{method}

\begin{figure*}[ht]
    \centering
    \includegraphics[width=17cm]{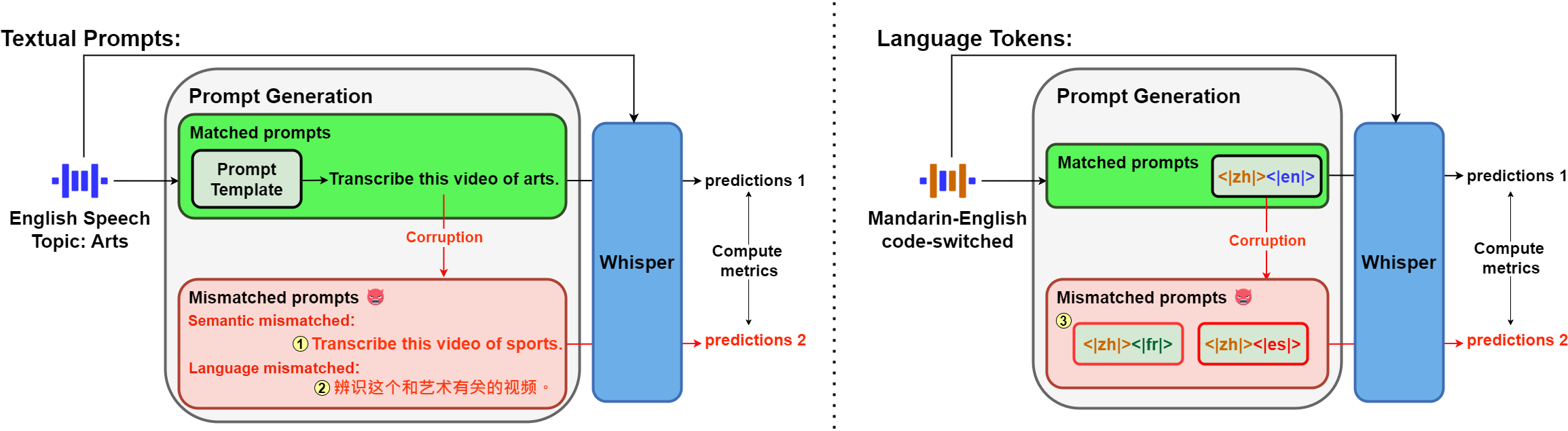}
    \caption{Illustration of assessing the understanding of textual prompts and language tokens. Matched prompts are first generated and then corrupted into mismatched prompts by changing the topic information, the prompt language (left), or one of the language tokens into the mismatched ones (right). The performances of Whisper with the matched and mismatched prompts are compared by metrics in Sec. \ref{metrics}.}
    \label{fig:corruption_example}
    \vspace{-10pt}
\end{figure*}

Our method is illustrated in Fig. \ref{fig:corruption_example}. We compare Whisper's performances based on conventional word error rate and the metrics detailed in Sec. \ref{metrics} in two contrasting scenarios: one with prompts containing correct and relevant information about the testing data, and the other with prompts corrupted by mismatched information. Ideally, if the model understands the prompts, performances in the former scenario should outperform the latter~\cite{webson-pavlick-2022-prompt, min-etal-2022-rethinking, shivagunde-etal-2024-deconstructing-context}. This comparative method assesses Whisper's prompt understanding, testing the model's ability to discern and utilize relevant information in prompts and offering a precise measure of prompt understanding.


Mismatched prompts are designed systematically to address potential confounding factors. Specifically, they are generated by introducing incorrect information that does not align with the testing data. For semantic understanding of textual prompts, we construct the prompt templates (Sec. \ref{template}) and introduce semantic corruption systematically (Sec. \ref{textualmethod}). The templates follow various patterns, ensuring our results are not biased by certain prompt structures. For the effect of textual prompt languages, the corruption is introduced by using prompts translated into a language different from the data while maintaining the meaning (Sec. \ref{prompt_language_method}), mitigating the effect of semantic variation. Regarding the language tokens, we provide a pair of language tokens of the languages occurring in the data and corrupt the pairs by replacing one token with a non-existent language (Sec. \ref{language_token_method}). No textual prompt is provided here and all the other settings are identical.

Although this work focuses on Whisper, our method can be generalized to other speech models. Different models may handle prompts and mismatches differently, but our systematic approach can still evaluate their prompt understanding capabilities. This generalizability ensures our method is a valuable tool for advancing prompt understanding research in speech-processing models.




\subsubsection{Prompt templates for textual prompts}
\label{template}
To explore Whisper's textual prompt understanding, we construct \textit{prompt templates} to generate textual prompts. These templates transform inputs into strings. For instance, a template $\mathtt{This}$ $\mathtt{utterance}$ $\mathtt{is}$ $\mathtt{about}$ $\{\mathtt{input}\}$ turns $\mathtt{arts}$ into a prompt ``This utterance is about arts". Using templates helps systematically generate textual prompts with the desired information encoded.

To ensure diversity, templates are first generated manually with various patterns, where some resemble those in prior works, categorized as follows with examples: (1) identity mappings, which use the input as prompts: $\{\mathtt{input}\}$; (2) task-informed instructions~\cite{wang2024can, lyricwhiz}: $\mathtt{Transcribe}$ $\mathtt{this}$ $\mathtt{video}$ $\mathtt{of}$ $\{\mathtt{input}\}$; (3) conversation-like templates~\cite{cookbook}: $\mathtt{So}$ $\mathtt{we}$ $\mathtt{were}$ $\mathtt{just}$ $\mathtt{talking}$ $\mathtt{about}$ $\{\mathtt{input}\}$; (4) indication of the input information in natural language form~\cite{yang2023investigating}: $\mathtt{This}$ $\mathtt{utterance}$ $\mathtt{is}$ $\mathtt{about}$ $\{\mathtt{input}\}$; (5) lists of input-related keywords~\cite{concat} generated by GPT-3.5~\cite{gpt}: $\mathtt{\{keyword\_1\},}$ $\mathtt{\{keyword\_2\}, ...,}$ $\mathtt{\{keyword\_N\}}$. In particular, human-generated templates of patterns (2), (3), and (4) are provided to GPT-3.5 to generate additional templates. All templates are generated in English. The resulting 10 templates generated by humans and GPT-3.5 are collected, and textual prompts are subsequently constructed with initial characters capitalized. For instance, given the input: $\mathtt{arts}$, the generated prompts of the aforementioned templates are (1) ``Arts", (2) ``Transcribe this video of arts", (3) ``So we were just talking about arts", (4) ``This utterance is about arts", and (5) ``Arts, culture, performing, visual".

\subsubsection{Semantic understanding of textual prompts}
\label{textualmethod}
We evaluate Whisper's semantic understanding of textual prompts using multi-domain monolingual ASR corpora, with each audio clip labeled with a specific topic. Multi-domain ASR is chosen to facilitate corruption by replacing topic labels in the prompts, enabling efficient generation of both matched and mismatched prompts.

As illustrated in Fig. \ref{fig:corruption_example}, we compare its performances when prompted by matched and mismatched topics. The prompts are generated using templates from Sec. \ref{template}, with topic labels as inputs, and translated to match the language of the testing data to prevent the impact of language inconsistency, discussed in Sec. \ref{prompt_language_method} and \ref{prompt_language}. This results in a set of English and Mandarin prompts. 

Since matched and mismatched prompts differ only in topic correctness, the performance gap between these scenarios indicates the model's sensitivity to topic information and measures its semantic understanding of textual prompts. We compare performance across subsets of each topic, examining prompts with correct and mismatched topics, using metrics from Sec. \ref{metrics}. In addition, the performances under the no-prompt settings are also discussed to gain more insights into the general effectiveness of prompting.

\subsubsection{Effect of textual prompt languages}
\label{prompt_language_method}
We explore whether the languages of textual prompts influence Whisper's performance, i.e., whether it gets enhanced by implicit language information in textual prompts. During the pre-training, the decoder predicts conditioned on the content of the previous speech segment, typically in the same languages as the testing segment. Consequently, textual prompts in the same language as the testing data should yield better performance since it is closer to the pre-training setting. To test this hypothesis, we compare Whisper's transcription accuracy on English and Mandarin data when prompted by English or Mandarin prompts with identical meanings.

To ensure semantic alignment between the Mandarin and English prompts, we use the prompts from Sec. \ref{textualmethod}, where the Mandarin prompts is obtained by translating English ones into Mandarin. This minimizes the impact of semantic differences and isolates the effect of language differences in the prompts. Both sets of prompts are applied to the testing corpora, and performances are compared using the metrics in Sec. \ref{metrics}, which reflect the model's ability to utilize language information from the textual prompts.

\subsubsection{Investigating the understanding of language tokens}
\label{language_token_method}
As prior works~\cite{concat, aditya2024attention} have shown that Whisper can be generalized to out-of-domain code-switching scenarios and unseen multilingual tasks through manipulations on the language tokens and the associated attention weights, it is worth investigating if Whisper understands the language information encoded in these tokens. Thus, we incorporate the understanding of the encoded information in the language tokens as part of Whisper's prompt understanding.

We use Mandarin-English and French-English code-switched ASR as probing tasks. The evaluation method is similar to that for textual prompts. During testing, the language tokens of two languages are concatenated~\cite{concat}, e.g., $<$$|\mathtt{zh}|$$>$$<$$|\mathtt{en}|$$>$, and provided to the model, where either both languages exist in the testing corpora or one of them is absent. The latter serves as the corrupted prompts. For example, when testing on Mandarin-English corpora, the provided language token pairs may be $<$$|\mathtt{zh}|$$>$$<$$|\mathtt{en}|$$>$, where both languages exist in the corpora, or pairs like $<$$|\mathtt{zh}|$$>$$<$$|\mathtt{fr}|$$>$, where one of them, French in this example, does not exist. However, pairs like $<$$|\mathtt{fr}|$$>$$<$$|\mathtt{es}|$$>$ will never be provided since both French and Spanish are absent in the corpora.

Code-switched ASR is chosen over monolingual ASR because only a language token will be provided for the latter, and providing the model with the wrong language token for monolingual ASR leads to speech-to-text translation into that language~\cite{concat}, making the analysis unfair. This is also why we only provide one mismatched language token instead of two. We compare Whisper's mixed error rates (MERs) when provided with entirely correct language token pairs versus partially correct ones. The difference in performances demonstrates Whisper's understanding of the encoded information in the language tokens.

\vspace{-5pt}
\subsection{Metrics}
\label{metrics}
\subsubsection{Word error rate (WER) and mixed error rate (MER)}
\label{mer}
The word error rate (WER) is employed for monolingual ASR. The mixed error rate (MER), defined as the WER with each Mandarin character treated as a word, serves as the metric for Mandarin-English and French-English CS ASR. The MER for French-English CS ASR aligns with conventional WER. However, for Mandarin-English CS ASR, pre-processing on ground truths and model outputs is necessary. This includes inserting spaces to treat Mandarin characters as words, standardizing Mandarin parts to simplified Mandarin, lowercasing non-Mandarin parts, and removing punctuation. The resulting MERs are computed and analyzed.

\subsubsection{Performance (PERF) and best performance (BPERF)}
\label{performance}
\begin{figure}[t]
    \centering
    \includegraphics[width=7.5cm]{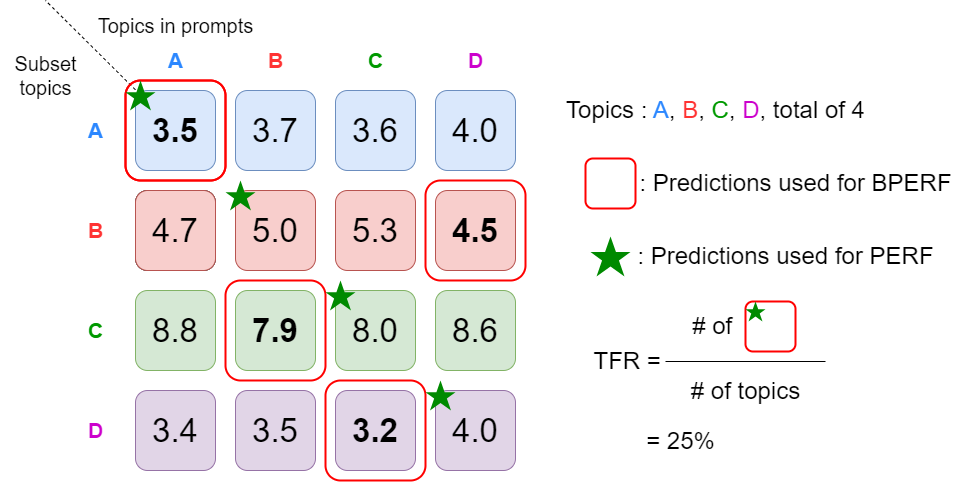}
    \setlength{\belowcaptionskip}{-15pt}
    \caption{
    Illustration of PERF, BPERF, and TFR. Red squares mark the minimum WER on each subset across all prompts. Green stars mark the WER on each subset when prompted by matched prompts.} 
    \label{fig:metrics}
\end{figure}

We define two metrics to quantitatively evaluate the impact of mismatched topic information: \textit{performance} (PERF) and \textit{best performance} (BPERF) of a prompt template.

To calculate these metrics, we first divide the testing set into subsets based on topic labels and generate textual prompts using the template with all topics. For each combination of subsets and prompts, we gather predictions and calculate the word error rate (WER). These WERs are organized into a matrix, exemplified by Fig. \ref{fig:metrics}, with subset topics as rows and prompt topics as columns.

The \textbf{performance (PERF)} of a template is the model's WER when entirely prompted with \textit{matched} topics via the template. This corresponds to the overall WER of predictions associated with the diagonal of the matrix. PERF assesses the model when prompted by purely matched information through a specific template.

Conversely, the \textbf{best performance (BPERF)} of a template assesses the lowest achievable WER when providing various topic information via the template. For each subset, all prompts are used, and the corresponding predictions with the lowest WER, exemplified by elements in red squares in Fig. \ref{fig:metrics}, are gathered. The BPERF is defined as the overall WER of the gathered predictions. The average PERF/BPERF is the average of PERFs/BPERFs across all templates. Comparing the average PERF and the average BPERF measures the impact of mismatched information on improvement or degradation. 

Though a single WER value may not fully demonstrate performances on semantic-related words and prompt understanding, comparing PERF and BPERF is still sufficient as a measurement since non-semantic errors like function words do not significantly vary across different prompts, which is empirically verified. Hence, the effect of non-semantic errors is canceled when comparing PERF and BPERF, and their difference demonstrates prompt understanding.

\subsubsection{Topic-following rate (TFR)}
\label{tfr}
Ideally, a model that comprehends textual prompts well should perform the best in a subset of a topic when prompted with that topic. This property is quantified by the \textit{topic-following rate} (TFR).

The TFR of a template is defined as follows: Let $n$ be the number of topics in the dataset and $f$ be the number of subsets where the model achieves the lowest WER when prompted with the topic-matched prompts. In other words, $f$ represents the number of rows whose minimum occurs on the diagonal of the matrix mentioned in Sec. \ref{performance}.
The TFR of the template on this dataset is $f/n$. For example, in Fig. \ref{fig:metrics}, where $n$ is 4 and $f$ is 1, resulting in a TFR of 25\%. The average TFR is the mean TFR of all templates.

While comparing PERF and BPERF in Sec. \ref{performance} quantifies the model's performance gap when prompted with matched versus mismatched information, TFR measures its overall sensitivity and adherence to the topic information. These metrics complement each other and are essential for quantitatively assessing the model's understanding of textual prompts.


\section{Experimental setups}
\subsection{Model}
We employ \textbf{Whisper-large-v3}, the latest version of Whisper trained on 5M hours of multilingual data with 1.5B parameters, referred to as Whisper henceforth. Greedy decoding is applied.

\subsection{Datasets}
\subsubsection{Multi-domain ASR corpora}
Multi-domain ASR corpora with associated topic labels are employed to investigate Whisper's understanding of textual prompts. We incorporate \textbf{GigaSpeech}~\cite{gigaspeech}, a multi-domain English ASR corpus. We utilize the training split\footnote{No models trained. Using training split for evaluation is acceptable. Though it is unknown whether these data were used during pre-training, the boost from pre-training can be canceled when we only focus on the difference of performances in two scenarios, since such boost will affect them similarly.} of its S subset, as the testing split lacks topic labels. Notably, we notice that some topics in GigaSpeech, e.g., ``People and Blogs" and ``Entertainment," are quite rough, and their data may be semantically related to other topics simultaneously. To avoid potential overlap, we select subsets of four highly distinct topics: ``Arts," ``Science and Technology," ``Nonprofits and Activism," and ``Sports," totaling 34.4 hours.

In addition to GigaSpeech, \textbf{ASCEND}~\cite{ascend}, a well-known multi-domain Mandarin-English code-switched corpus, is also adopted. It comprises conversational speech with diverse topics: ``education", ``technology", ``persona", ``philosophy", and ``sports". For experiments about textual prompts, to prevent potential bias caused by code-switching~\cite{yang2023investigating}, which is not the focus of these experiments, we utilize data entirely in Mandarin or English from the training split, creating subsets denoted as \textbf{ASCEND-zh} and \textbf{ASCEND-en}, with durations of 3.5 and 1.7 hours, respectively. The training split is used since only that split contains all the aforementioned topics.

\subsubsection{Code-switched ASR corpora}
To examine Whisper's understanding of the encoded information in language tokens, several code-switched ASR corpora are included in our experiments. The entire testing split of the aforementioned ASCEND is adopted again, totaling 0.92 hours. Additionally, we adopt \textbf{CSZS-correct-zh} and \textbf{CSZS-correct-fr}, which are Mandarin-English and French-English code-switched datasets~\cite{cszs} featuring high-quality intra-sentential code-switched speech. The testing splits are adopted, totaling 4.1 and 15.4 hours, respectively.

\section{Results and discussion}
We discuss our results. The 95\% confidence intervals (CIs) obtained from bootstrapping~\cite{bootstrap}, with 1k bootstrap sets, are shown in brackets in the tables. CIs for TFRs are omitted due to the limited number of templates. Each experiment takes around 10 hours on a V100.

\subsection{Semantic understanding of textual prompts}
\label{textual}

We examine Whisper's semantic understanding of textual prompts by contrasting its performances when provided with correct versus mismatched topics. To mitigate potential biases from language discrepancies between the textual prompts and the speech, the prompts are translated into the data language. Results of Whisper when not prompted are also included. Our results are in Table \ref{tab:tfrs}.

\begin{table}[ht]
\setlength\tabcolsep{2.5 pt}
\renewcommand{\arraystretch}{0.1}

\caption{Average TFR(\%), PERF(\%), BPERF(\%) of Whisper. ``Relative improvement" is the relative improvement of BPERF to PERF, representing the highest relative gain of using mismatched topics over matched ones for the employed templates. ``no-prompt" denotes Whisper's performance when not prompted.}
\vspace{-5pt}
\centering


\resizebox{\columnwidth}{!}{
\begin{tabular}{cccccc}

\toprule
Dataset & \makecell[c]{Avg\\TFR ($\uparrow$)}  & \makecell[c]{Avg \\PERF ($\downarrow$)} & \makecell[c]{Avg\\BPERF ($\downarrow$)} & \makecell[c]{Relative \\ improvement} & no-prompt\\
\midrule
\midrule


ASCEND-zh & 22.0\% & \makecell[c]{15.63\\$[$15.61, 15.68$]$} & \makecell[c]{14.92\\$[$14.90, 14.97$]$} & 5\% & \makecell[c]{12.21\\$[$11.80, 12.68$]$} \\
\midrule
ASCEND-en & 30.0\% & \makecell[c]{23.55\\$[$23.54, 23.60$]$} & \makecell[c]{22.69\\$[$22.69, 22.70$]$} & 4\% & \makecell[c]{22.58\\$[$21.63, 23.43$]$}\\
\midrule
GigaSpeech & 17.5\% & \makecell[c]{5.50\\$[$5.48, 5.52$]$} & \makecell[c]{4.90\\$[$4.89, 4.90$]$} & 11\% & \makecell[c]{4.28\\$[$4.18, 4.38$]$} \\






\bottomrule

\end{tabular}
}
\label{tab:tfrs}
\end{table}

From Table \ref{tab:tfrs}, Whisper's average topic-following rates (TFRs) are markedly unsatisfactory on the testing corpora, typically below 30\%. This indicates that \textbf{Whisper does not perform better when prompted by matched information in general}.

The PERFs and BPERFs of Whisper on the evaluation corpora are analyzed. First of all, we notice that \textbf{Whisper's average PERFs/BPERFs are typically worse than the no-prompt settings}. We further compare the PERFs and BPERFs. To ensure the comparison between PERFs and BPERFs can reflect the model's semantic prompt understanding, we first compute the word error rates for non-semantic errors like function words and observe a maximum difference of only 0.5\% between the predictions contributing to PERFs and BPERFs. This negligible difference suggests that such errors occur almost equally in settings with both matched and mismatched prompts, supporting our hypothesis in Section \ref{performance}. Thus, the performance discrepancies between BPERFs and PERFs effectively indicate the semantic understanding of textual prompts. From Table \ref{tab:tfrs}, the mismatched information can induce relative improvements of BPERF to PERF of up to 11\%, which are highly significant. Such improvements are commonly observed for templates inducing better or worse performances than the no-prompt baseline, showing that besides low TFRs, \textbf{Whisper gets greatly enhanced by irrelevant and incorrect information in textual prompts}. Given that there are already 10 diverse templates resembling prior works in our experiments, which is sufficient to mitigate the potential bias of the template choices, this result indicates that \textbf{textual prompts that make sense to humans do not always bring improvements}.

We also explore the relationship between Whisper's TFR and PERFs/BPERFs across diverse prompt templates to gain a deeper understanding of its behavior. High TFR and good PERF/BPERF for a template indicate the model's effective focus on topic information and transcription accuracy when using that template, respectively. By intuition, templates with higher TFRs should correspond to better (lower) PERFs/BPERFs, as they prompt the model to consider topic information more accurately. We examine this hypothesis.

\begin{figure}[t]
    \centering
    \includegraphics[width=8cm]{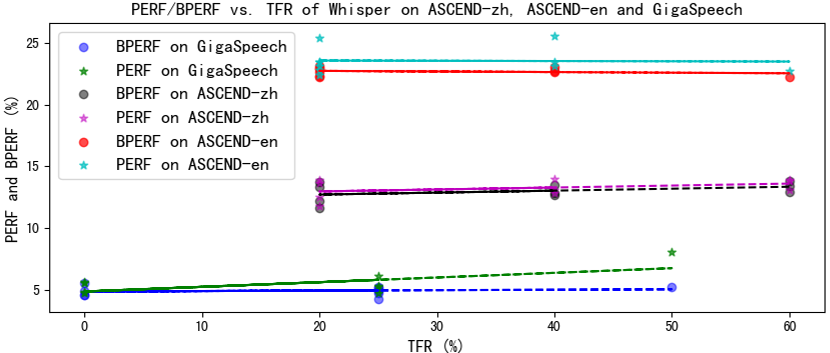}
    \vspace{-2mm}
    \setlength{\belowcaptionskip}{-15pt}
    \caption{Linear regression of PERF/BPERF of templates on the corresponding TFR. Points in the figure represent the PERF/BPERF and TFR prompted with specific templates.}
    \label{fig:regression}
\end{figure}

Linear regression is conducted and presented in Fig. \ref{fig:regression}. Interestingly, the results contradict our hypothesis. The hypothesis suggests negative slopes of the regression since lower PERF/BPERF values indicate better transcription quality. However, near-to-zero or even positive slopes of the regressions are observed. In general, templates guiding Whisper to understand topic information better, i.e., with higher TFRs, do not simultaneously enhance transcribing accuracy, i.e., lower PERF/BPERF. This indicates that \textbf{high TFR and good PERF/BPERF (good transcribing accuracy) of prompt templates are not positively correlated for Whisper}, which is another counter-intuitive behavior besides the previous observations. 

These counter-intuitive behaviors suggest that \textbf{Whisper may not understand textual prompts in a human-expected way}. Prompts that seem informative to humans may not be instructive to Whisper. It may process and comprehend prompts in a non-human manner. For instance, prompts that enhance Whisper's performance might follow patterns beyond human perception and semantic meaningfulness, which activate and improve Whisper. We believe how to automatically find effective prompts and establish human-like semantic understanding for Whisper will be essential for developing universal speech models. We take these as our future work.



\subsection{Effect of textual prompt languages}
\label{prompt_language}
We discuss the effect of the language of the textual prompts. The results of providing English and Mandarin prompts are in Table \ref{tab:prompt_language}.
\begin{table}[ht]\scriptsize
\setlength\tabcolsep{1 pt}
\renewcommand{\arraystretch}{0.2}

\caption{Average TFR(\%), PERF(\%), and BPERF(\%) of Whisper on ASCEND-zh, ASCEND-en, and GigaSpeech when prompted by prompts in Mandarin (zh) and English (en).}
\centering
\vspace{-5pt}

\resizebox{\columnwidth}{!}{
\begin{tabular}{c|cc|cc|cc}
\toprule
 & \multicolumn{2}{c|}{Avg TFR ($\uparrow$)} & \multicolumn{2}{c|}{Avg PERF ($\downarrow$)} & \multicolumn{2}{c}{Avg BPERF ($\downarrow$)} \\
\midrule
Prompt language & \makecell[c]{zh} & \makecell[c]{en}&\makecell[c]{zh} &\makecell[c]{en} & \makecell[c]{zh} & \makecell[c]{en} \\
\midrule
\midrule

ASCEND-zh & 22.0\% & \textbf{38.0\%} & \makecell[c]{15.63\\$[$15.61,15.68$]$} & \makecell[c]{\textbf{13.24}\\$[$13.24,13.27$]$} & \makecell[c]{14.92\\$[$14.90,14.97$]$} & \makecell[c]{\textbf{12.99}\\$[$12.98,13.01$]$} \\
\midrule

ASCEND-en & 14.0\% & \textbf{30.0\%} & \makecell[c]{27.93\\$[$27.81,28.16$]$} & \makecell[c]{\textbf{23.55}\\$[$23.54,23.60$]$} & \makecell[c]{26.09\\$[$25.98,26.24$]$} &\makecell[c]{\textbf{22.69}\\$[$22.69,22.71$]$} \\
\midrule

GigaSpeech & \textbf{22.5\%} & 17.5\% & \makecell[c]{7.79\\$[$7.74,7.84$]$} & \makecell[c]{\textbf{5.50}\\$[$5.48,5.52$]$} & \makecell[c]{6.23\\$[$6.19,6.27$]$} & \makecell[c]{\textbf{4.90}\\$[$4.89,4.90$]$} \\

\bottomrule
\end{tabular}
}
\label{tab:prompt_language}
\vspace{-15pt}
\end{table}

Surprisingly, it is observed that, in most cases, \textbf{Whisper performs better when prompted in English compared to Mandarin, even on Mandarin data}, despite the prompts having identical meanings. The performance gaps caused by using English prompts on Mandarin data are significant, with a maximum relative improvement of 15\% for PERF and 13\% for BPERF on ASCEND-zh. This contradicts the hypothesis in Sec. \ref{prompt_language_method} that the model would perform better when prompted in the same language as the testing data, though this setting is much closer to the pre-training scenarios than that has prompts and testing speech in different languages.

This might be attributed to the substantial disparity in the amount of training data for English and Mandarin~\cite{whisper}. Whisper was primarily trained on English data, and the previous context on which the generation was conditioned was typically in English. This might make Whisper more familiar with English than other languages. Thus, when replacing the previous context with textual prompts, English prompts may elicit better performance than their Mandarin counterparts, regardless of the language discrepancy between prompts and the speech.



\subsection{Encoded information in language tokens}
\label{language_token}

As the prompting methods with language tokens are well-studied compared with those with textual prompts~\cite{concat, yang2023investigating}, we mainly focus on the effect of misleading language tokens, which is not covered by prior works, to gain more insights. Our results are shown in Tables \ref{tab:zhen} and \ref{tab:fren}, where $<$$|\mathtt{zh}|$$>$, $<$$|\mathtt{en}|$$>$, $<$$|\mathtt{es}|$$>$, $<$$|\mathtt{fr}|$$>$, and $<$$|\mathtt{it}|$$>$ represents Mandarin, English, Spanish, French, and Italian, respectively. The numbers of Mandarin and French words are counted with an off-the-shelf language identification tool~\cite{lingua}.
\begin{table}[ht]\scriptsize
\setlength\tabcolsep{1.5 pt}
\renewcommand{\arraystretch}{0.2}

\caption{MER (\%) on CSZS-correct-zh and ASCEND with different combinations of language tokens. The ``zh word count" is the number of generated Chinese words on CSZS-correct/ASCEND. The ground truth of the example is \begin{CJK}{UTF8}{gbsn} ``Let's 趁机 take some pictures". \end{CJK}} 
\centering
\vspace{-5pt}
\resizebox{\columnwidth}{!}{
\begin{tabular}{ccccc}
\toprule
Language tokens & CSZS-correct-zh & ASCEND & \makecell[c]{zh word count}& \makecell[c]{Prediction\\ examples}\\
\midrule
\midrule
$<$$|\mathtt{zh}|$$>$$<$$|\mathtt{en}|$$>$ & \makecell[c]{\textbf{26.76} \\$[$22.35, 31.72$]$} & \makecell[c]{\textbf{21.93} \\$[$17.21, 29.92$]$} & 29491 / 9353 &  \begin{CJK}{UTF8}{gbsn} \makecell[c]{Let's Genji take \\some pictures
} \end{CJK} \\
\midrule
$<$$|\mathtt{zh}|$$>$$<$$|\mathtt{es}|$$>$ & \makecell[c]{54.24\\$[$48.75, 60.23$]$} & \makecell[c]{26.51\\$[$24.39, 28.68$]$} & 44334 / 10253& \begin{CJK}{UTF8}{gbsn} \makecell[c]{让我们来拍\\一些照片} \end{CJK} \\
\midrule
$<$$|\mathtt{zh}|$$>$$<$$|\mathtt{fr}|$$>$ & \makecell[c]{58.85\\$[$52.69, 65.78$]$} & \makecell[c]{26.59\\$[$24.46, 28.72$]$} & 46729 / 10242& \begin{CJK}{UTF8}{gbsn} \makecell[c]{让我们来拍\\一些照片} \end{CJK} \\
\midrule
$<$$|\mathtt{zh}|$$>$$<$$|\mathtt{it}|$$>$ & \makecell[c]{50.05\\$[$44.39, 55.50$]$} & \makecell[c]{30.73\\$[$25.22, 39.81$]$} & 42040 / 10491& \begin{CJK}{UTF8}{gbsn} \makecell[c]{让我们来拍\\一些照片} \end{CJK} \\
\bottomrule
\end{tabular}
}
\label{tab:zhen}
\end{table}

\begin{table}[ht]\scriptsize
\setlength\tabcolsep{1.5 pt}
\renewcommand{\arraystretch}{0.2}

\caption{MER (\%) on CSZS-correct-fr with various oairs of language tokens. The ``fr word counts" is the number of generated French words. The example ground truth is ``He is a member of the Discovery Institute un de la droite religieuse of the American right".} 
\centering
\vspace{-5pt}
\resizebox{\columnwidth}{!}{
\begin{tabular}{cccc}
\toprule
Language tokens & CSZS-correct-fr & fr word counts & Prediction examples\\
\midrule
\midrule
$<$$|\mathtt{fr}|$$>$$<$$|\mathtt{en}|$$>$ & \makecell[c]{\textbf{30.55}\\$[$29.74, 31.67$]$} & 98467 & \makecell[c]{He is a member of the \\Discovery Institute, and a \\member of the American Right.} \\
\midrule
$<$$|\mathtt{fr}|$$>$$<$$|\mathtt{es}|$$>$ & \makecell[c]{36.31\\$[$35.46, 37.36$]$} & 107071 & \makecell[c]{Il est le même membre de l'Institut \\ de la Découverte, un de la droite \\ religieuse de l'Amérique.} \\
\midrule
$<$$|\mathtt{fr}|$$>$$<$$|\mathtt{it}|$$>$ & \makecell[c]{34.07\\$[$33.31, 34.99$]$} & 102229 & \makecell[c]{Il même membre de l'Institut \\ Discovery, un des droits religieux \\de l'Amérique des Nations Unies.} \\
\midrule
$<$$|\mathtt{fr}|$$>$$<$$|\mathtt{zh}|$$>$ & \makecell[c]{32.71\\$[$31.86, 33.93$]$} & 100840 & \makecell[c]{Il est le même membre de l'Institut \\ de la Découverte, un de la droite \\ religieuse de l'Amérique.} \\
\bottomrule
\end{tabular}
}
\label{tab:fren}
\vspace{-10pt}
\end{table}

On average, the model performs the best with entirely correct language tokens. The relative performance degradation resulting from wrong language tokens is significant, up to 120\% for CSZS-correct-zh, 40\% for ASCEND, and 19\% for CSZS-correct-fr. 

From Yang et al.~\cite{yang2023investigating}, Whisper tends to generate predictions exclusively in or translate parts of speech into the dominant language of the utterances in CS ASR, exemplified in the initial rows of Table \ref{tab:zhen} and \ref{tab:fren}. Notably, when given partially correct language tokens (one existing and one non-existing language), Whisper generates predictions in the existing languages rather than English or the non-existing languages, despite the dominant language being English. Examples are provided in Table \ref{tab:zhen} and \ref{tab:fren}. It is quantitatively supported by the word counts, where significantly more Mandarin or French words are generated with partially correct language tokens.


We interpret this phenomenon as follows: When presented with a pair of language tokens, Whisper generates predictions based on mixed information from these tokens, guided by speech information from the encoder to decide the language of the current token. The model dynamically adjusts its focus on the two language tokens to predict a code-switched token sequence for code-switched speech. If one of the provided language tokens is nonexistent in the speech, the encoder's guidance is likely to steer the model to focus on the correct language token and ignore the incorrect one since there is no speech information about the latter, resulting in the ability to ignore the incorrect language token. This ability can be attributed to multi-task pre-training encompassing language identification and ASR. Our interpretation can reasonably explain the phenomenon observed in Table \ref{tab:zhen} and \ref{tab:fren} and can be applied to interpret prior works that manipulate Whisper's special tokens to change its behavior~\cite{concat}. In sum, \textbf{Whisper can prioritize the relevant language information and ignore the irrelevant one in language tokens}.

\section{Conclusion and Future Work}
We study Whisper's prompt understanding. Despite the widespread use of prompting Whisper, we discovered that informative prompts in human perception do not improve Whisper in general, and better performance is not guaranteed even when demonstrating better comprehension given a prompt template. We found that English prompts yield better performance than Mandarin counterparts on both English and Mandarin data, possibly due to differences in the training data volume for these languages. It was also noted that Whisper can filter out misleading information in language tokens, likely due to extensive pre-training on language identification and ASR. Future work includes automatically constructing effective prompts and establishing speech models with genuine prompt understanding. We encourage the community to carefully revisit Whisper's prompting methods and provide more investigations.

\section{Acknowledgement}
We acknowledge the computational and storage support from National Center for High-performance Computing (NCHC) of National Applied Research Laboratories (NARLabs) in Taiwan.
\bibliographystyle{IEEEbib}
\bibliography{refs}

\end{document}